\def\year{2021}\relax
\documentclass[letterpaper]{article} 
\usepackage{aaai21}  
\usepackage{times}  
\usepackage{helvet} 
\usepackage{courier}  
\usepackage[hyphens]{url}  
\usepackage{graphicx} 
\usepackage{natbib}  
\usepackage{caption} 
\usepackage{amsmath}
\usepackage{amssymb}
\usepackage{amsfonts}
\usepackage{enumitem}
\usepackage{graphicx}
\usepackage{xcolor}
\graphicspath{{Figure/}}
\title{Learning to Select External Knowledge with Multi-Scale Negative Sampling}
\author{Huang He\thanks{~First two authors contributed equally to this work.}, Hua Lu\footnotemark[1]\thanks{~Work was done during internship at Baidu.}, Siqi Bao, Fan Wang, Hua Wu, Zhengyu Niu, Haifeng Wang\\}
\affiliations{Baidu Inc., China\\
\{hehuang, v\_luhua01, baosiqi, wang.fan, wu\_hua, niuzhengyu, wanghaifeng\}@baidu.com}

\begin{document}

\maketitle

\begin{abstract}
The Track-1 of DSTC9 aims to effectively answer user requests or questions during task-oriented dialogues, which are out of the scope of APIs/DB. By leveraging external knowledge resources, relevant information can be retrieved and encoded into the response generation for these out-of-API-coverage queries. In this work, we have explored several advanced techniques to enhance the utilization of external knowledge and boost the quality of response generation, including \textit{schema guided knowledge decision}, \textit{negatives enhanced knowledge selection}, and \textit{knowledge grounded response generation}. To evaluate the performance of our proposed method, comprehensive experiments have been carried out on the publicly available dataset. Our approach was ranked as the best in human evaluation of DSTC9 Track-1.
\end{abstract}

\section{Introduction}
Task-oriented dialogue agents have been widely used in our daily lives, such as digital personal assistant, customer service bot, and so on. Given that these agents generally rely on pre-defined APIs to provide services, they cannot handle user queries beyond the API's coverage. For instance, booking a seat at a restaurant can be handled with a pre-defined API. However, requesting the noise level at this restaurant might go beyond the API's coverage and lead to the failure of this system. Under the circumstances, users need to find out the information themselves, by visiting the website description, customer reviews or FAQs. In fact, for most out-of-API-coverage queries, relevant information might already exist in external resources. 

To tackle the above problem, external knowledge is extracted and employed to boost the capacity of task-oriented dialogue system \cite{kim2020beyond}. To this end, an augmented dataset with external knowledge access is constructed based on MultiWOZ 2.1 \cite{eric2020multiwoz}. The conversation in MultiWOZ 2.1 is about touristic information seeking between a tourist and a clerk, confined by pre-defined APIs. In the augmented dataset, out-of-API-coverage utterances with external knowledge access are inserted accordingly into the original conversation.

Distinct with open-domain conversations \cite{dinan2018wizard, lian2019learning, fan2020augmenting}, task-oriented conversation needs to deliver information accurately to satisfy user's needs. Therefore, it encounters more stringent requirements on knowledge selection and utilization. First, the system has to pick out the most accurate knowledge snippet from the large external database, not just relevant ones. For example, to respond to the user with the opening hours of a particular museum, the references from other museums are not very useful. Second, to generate an accurate and coherent response, the system needs to carry out elaborate processing and reasoning with the retrieved knowledge snippet and dialogue context. 

In DSTC9 Track-1, the retrieval-augmented response generation has been split into three successive tasks. First, given the dialogue context, the system decides whether to trigger external knowledge access or not. Second, for the turns requiring external knowledge, the system selects the most appropriate knowledge snippets. Third, the system generates the responses given the dialogue context and selected knowledge. In this work, to enhance task-oriented dialogue generation, a complete solution is introduced to the three tasks. In Task1, we propose the \textit{schema guided knowledge decision}, which takes the functions of APIs and external knowledge into consideration. In Task2, we introduce the \textit{negatives enhanced knowledge selection}, which includes multi-scale negatives to increase the training difficulty and boost the selection performance. In Task3, to obtain coherent and accurate responses, we leverage powerful pre-trained models for \textit{knowledge grounded response generation}. To evaluate the performance of our proposed solution, comprehensive experiments have been carried out on the publicly available dataset. Our approach was ranked as the best in human evaluation of DSTC9 Track-1.

\section{Methodology}
In this section, we will discuss the following strategies in detail, including schema guided knowledge decision, negatives enhanced knowledge selection, and knowledge grounded response generation. 

\subsection{Schema Guided Knowledge Decision}
To determine whether to seek external knowledge or not, the most straightforward way is to rely on the dialogue context \cite{kim2020beyond}. However, such an approach typically captures the frequent keywords or semantic patterns in the context, which might lead to biased decision and suffer from poor performance on the unseen domain/locale conversations. In practice, the system is supposed to know the functions of APIs and external knowledge before making a choice between them. For the sake of a comprehensive decision, we take the dialogue context as well as the API/knowledge functions into consideration. 

\begin{figure}
	\centering
	\includegraphics[width=0.38\textwidth]{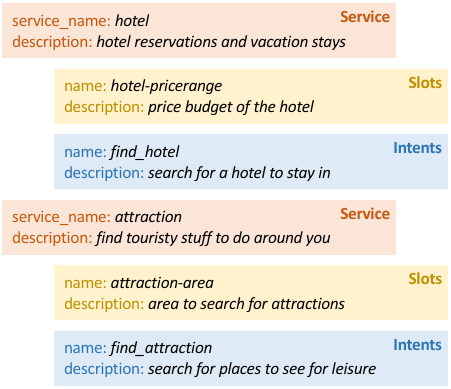}
	\caption{Some schema descriptions from MultiWOZ 2.2.}
	\label{fig:schema}
\end{figure}

While it is challenging to encode the structured APIs into the decision making process. Inspired by the recent progress in schema description \cite{shah2019robust, eric2020multiwoz, rastogi2020schema}, we employ the natural language descriptions to represent the functions of APIs. Some schema descriptions from MultiWOZ 2.2 \cite{zang2020multiwoz} are illustrated in Figure \ref{fig:schema}, with supported slots and intents listed under each service/API. The schema descriptions are denoted as $S=\{s_1, s_2, \cdots, s_m\}$, where $s_i$ is one slot/intent description from MultiWOZ 2.2. The external knowledge snippets are represented as $K=\{k_1, k_2, \cdots, k_n\}$, where $k_i$ represent the $i$-th knowledge snippet. The dialogue context is referred as $C_t=\{u_1, u_2, \cdots, u_t\}$, where $u_i$ is the $i$-th utterance in a multi-turn conversation and $t$ is the current time step. In the schema guided knowledge decision, we will estimate the following probability $p_{\text{decision}}(l_x=1|C_t,x)$, where $x$ can be one schema description $s_i$ or knowledge snippet $k_i$. $l_x$ stands for the label to choose $x$ or not given the dialogue context. The input is fed into transformer network in the following format: [CLS] $C_t$ [SEP] $x$ [SEP], and the hidden embedding of [CLS] in the last layer is used to estimate the above probability. 

During the training process, mixed samples from schema descriptions and knowledge snippets are combined together and learned to minimize the following loss:
\begin{equation}
\begin{split}
    \mathcal{L}_{\text{decision}}=&-\sum_i\log p_{\text{decision}}(l_{x_i}=1|C_t,x_i)\\
    &-\sum_j\log p_{\text{decision}}(l_{x_j^-}=0|C_t,x_j^-)
\end{split}
\end{equation}
For the conversational turns that rely on API services, positive samples $x_i$ are collected with corresponding slot/intent descriptions, and negative samples $x_j^-$ include other schema descriptions and knowledge snippets. For the conversational turns that require external knowledge access, positive samples $x_i$ are collected with corresponding knowledge snippets, and negative samples $x_j^-$ include other snippets and schema descriptions. 

During inference, the knowledge-seeking turn is determined as follows. If the following condition is met:
\begin{equation}\nonumber
    \max\limits_{k_i\in K}~p_{\text{decision}}(l_{k_i}=1|C_t,k_i) \ge \max\limits_{s_i\in S}~p_{\text{decision}}(l_{s_i}=1|C_t,s_i)
\end{equation}
the system will consult the external knowledge snippets for subsequent response generation. Otherwise, the system will keep relying on the API services for response generation. It is notable that besides knowledge-seeking turn detection, the above estimated probability $\max\limits_{k_i\in K}~p_{\text{decision}}(l_{k_i}=1|C_t,k_i)$ is also passable for knowledge selection. 

\subsection{Negatives Enhanced Knowledge Selection}
Once determined to trigger the external knowledge access, the next move is to select the appropriate knowledge snippet. Within this section, we will elaborate more on the relevance estimation between each knowledge snippet and the dialogue context $p_{\text{selection}}(l_{k_i}=1|C_t,k_i)$.

Usually, the relevance function is trained to separate positive samples from those randomly selected negative samples. Given that the space of negative samples is extremely large, random sampling might lead to coarse-grained class separation, which is insufficient for fine-grained knowledge selection. Recently, it has been recognized that the selection of negative samples is crucial in boosting the capacity of retrieval systems \cite{henderson2017efficient, karpukhin2020dense}. In this task, we include multi-scale negatives to strengthen the ability of fine-grained relevance estimation. The training objective is to minimize the following loss:
\begin{equation}
\begin{split}
    \mathcal{L}_{\text{selection}}=&-\log p_{\text{selection}}(l_{k_i}=1|C_t,k_i)\\
    &-\sum_{j}\log p_{\text{selection}}(l_{k_{i,j}^-}=0|C_t,k_{i,j}^-)
\end{split}
\end{equation}

The negatives samples $k_{i,j}^-$ are collected from distinct scales to increase the training difficulty. (1) Random: one knowledge snippet is randomly selected from the whole set $K$; (2) In-Domain: one knowledge snippet is randomly selected from those within the same domain as the positive sample $k_i$; (3) In-Entity: one knowledge snippet is randomly selected from those belonging to the same entity as the positive sample. (4) Cross-Entity: one knowledge snippet is randomly selected from those belonging to the aforementioned entity in the dialogue context. The training difficulty increases along with the refinement of the negative sample's granularity. During training, the ratio of positive to negative training samples is 1:4. During inference, the optimal knowledge snippet can be selected in the following way:
\begin{equation}\label{eq:selection}
    k^*=\max\limits_{k_i\in K}~p_{\text{selection}}(l_{k_i}=1|C_t,k_i)
\end{equation}

\subsection{Knowledge Grounded Response Generation}
There are two basic requirements for knowledge grounded response generation. First, the generated response needs to be coherent with the dialogue context, remaining a smooth conversation flow. Second, the response needs to express the information accurately, without deviating from the original knowledge snippet. Although there are several options to produce the response, it is not easy to simultaneously meet these requirements. For instance, with an open-domain chatbot, it is capable to produce one coherent response without external knowledge, but incapable to provide necessary information towards the user. Another way is to forward the retrieved knowledge snippet directly as the reply, resulting in incoherence of the conversation flow. One compromise solution is to employ machine comprehension techniques to extract spans from the knowledge snippet as the response. However, it still might fail when the accurate answer is not contained within the surface contents. As such, it is challenging to utilize knowledge accurately and generate high-quality responses.
\begin{figure}
	\centering
	\includegraphics[width=0.48\textwidth]{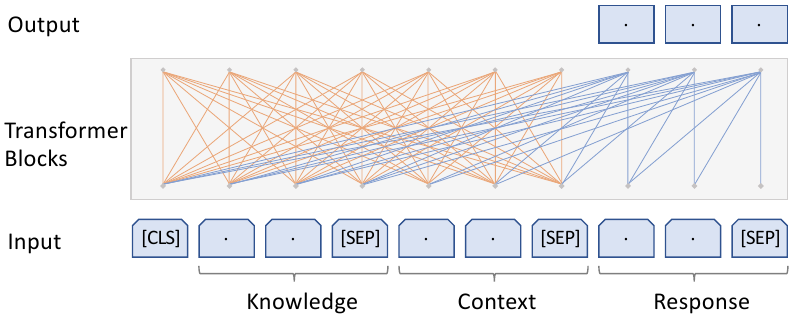}
	\caption{Knowledge grounded response generation. \textcolor{orange}{Orange} lines denote bi-directional attention, and \textcolor{blue}{blue} lines denote uni-directional attention}
	\label{fig:generation}
\end{figure} 

In this paper, we leverage powerful pre-trained models for knowledge grounded response generation. The network infrastructure is sketched in Figure \ref{fig:generation}. The backbone of the pre-training network consists of transformer blocks. The input to the network is the sum of the following four representations \cite{devlin2019bert, bao2019plato}.
\begin{itemize}
	\item Token Embedding. Following the conventional pre-processing, the input text is tokenized into byte-pair-encoding (BPE) tokens \cite{sennrich2016neural}. 
	
	\item Segment Embedding. To better differentiate the input information, distinct segment embeddings are assigned to the knowledge snippet, dialogue context and response.   
	
	\item Role Embedding. As the multi-turn conversation is interactive, role embeddings are employed to distinguish the utterances from different characters.
	
	\item Position Embedding. To obtain better extensibility on the input length, relative position embeddings are embraced in this generation network.
\end{itemize}	

As for the self-attention mechanism, bi-directional attention is enabled for better natural language understanding, and uni-directional attention is employed for auto-regressive response generation. The training objective is to minimize the negative log-likelihood (NLL) loss:
\begin{equation}
	\mathcal{L}_{\text{generation}} =-\mathbb{E} ~\log p_{\text{generation}}(r|C_t,k)
\end{equation}
where $r$ refers to the target response. During training, we utilize the golden knowledge snippet $\tilde{k}$ for response generation. During inference, we rely on the knowledge snippet $k^*$ retrieved with Equation \eqref{eq:selection} for response generation. 

\section{Experiments}
\subsection{Settings}
In DSTC9 Track-1, one augmented dataset \cite{kim2020beyond} is constructed based on MultiWOZ 2.1 \cite{eric2020multiwoz}. This dataset is about touristic information seeking between a tourist and a clerk. Besides the conventional API-related utterances, out-of-API-coverage utterances are inserted with external knowledge access. The detailed statistics of the augmented dataset are summarized in Table \ref{tab:dataset}. To evaluate the generalization ability of task-oriented dialogue systems, some unseen conversations from new domains or locales are included in the test set.
\begin{table}
	\centering
	\includegraphics[width=0.48\textwidth]{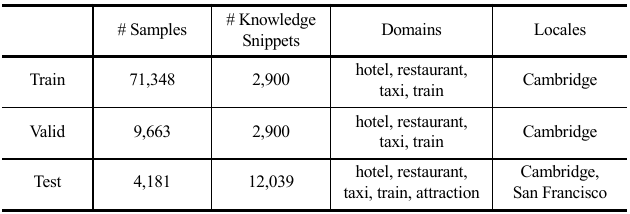}
	\caption{Dataset statistics of DSTC9 Track-1.}
	\label{tab:dataset}
\end{table}

\begin{table*}
	\centering
	\includegraphics[width=\textwidth]{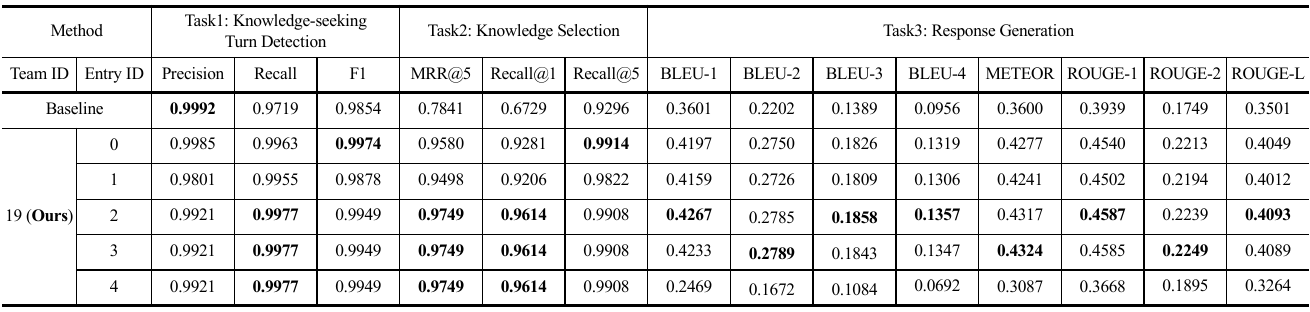}
	\caption{Experimental results on the validation set, with the highest value written in bold.}
	\label{tab:valid}
\end{table*} 
\begin{table*}
	\centering
	\includegraphics[width=\textwidth]{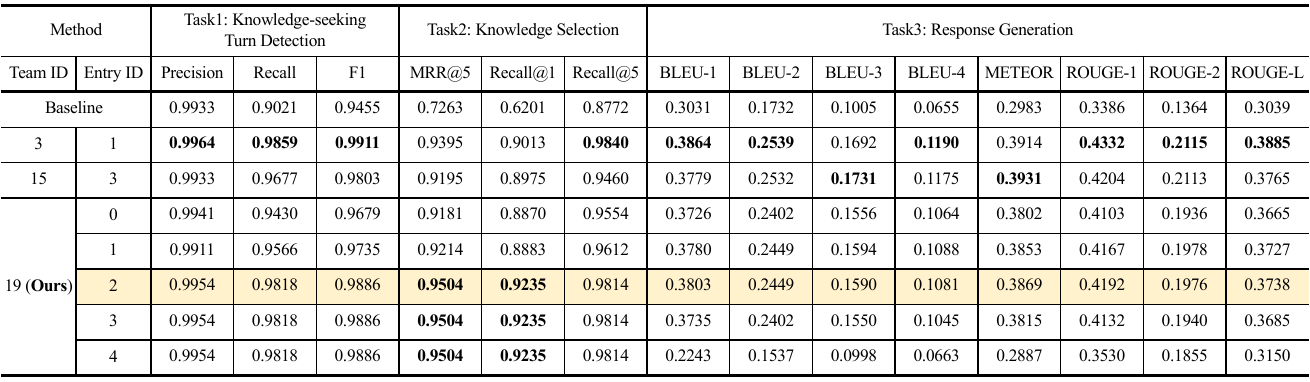}
	\caption{Experimental results on the test set, with the highest value written in bold.}
	\label{tab:test}
\end{table*} 
In DSTC9 Track-1, it focuses on the conversational utterances that require external knowledge access. The evaluation covers the following three successive tasks.
\begin{itemize}
\item Task1: Knowledge-seeking Turn Detection. The model needs to decide whether to trigger external knowledge access or not, given the dialogue context. The evaluation metrics of this task include precision, recall, and F1.

\item Task2: Knowledge Selection. For the conversational turns that require external knowledge access, the model needs to select appropriate knowledge snippets. The evaluation metrics of this task include mrr@5 \cite{voorhees1999trec}, recall@1, and recall@5.

\item Task3: Response Generation. The model needs to produce the responses given the dialogue context and selected knowledge snippets. The automatic evaluation metrics of this task include BLEU-1/2/3/4 \cite{papineni2002bleu}, Meteor \cite{denkowski2014meteor} and ROUGE-1/2/L \cite{linrouge}.
\end{itemize}

In the experiments, we leverage large-scale pre-training models to boost the performance of three tasks. The models are pre-trained with 684M (context, response) training samples extracted from Reddit. The vocabulary has 8k BPE subwords constructed with the SentencePiece library. The pre-training process is carried out via curriculum learning as PLATO-2 \cite{bao2020plato}. In the first stage, a generation model is trained to minimize the negative log-likelihood (NLL) loss. In the second stage, an evaluation model is further trained to minimize the sentence-order prediction (SOP) loss \cite{lan2019albert}. The evaluation model is used for the fine-tuning of Task1 and Task2. The generation model is employed for the fine-tuning of Task3. All the models have 32 transformer blocks and 32 attention heads, with the hidden embedding dimension of 2048. 

\subsection{Experimental Results}
During the competition, each team is able to submit at most five entries. The methods that we use in each entry are described as follows:
\begin{itemize}
\item Entry 0. In Task1, the knowledge-seeking turn detection is estimated based on the dialogue context. In Task2, the knowledge selection is enhanced with multi-scale negatives training. In Task3, the response is generated using beam search, with a beam size of 5.

\item Entry 1. In Task1, the proposed schema guided knowledge decision is adopted for knowledge-seeking turn detection. The settings of Task2 and Task3 are the same as entry 0. 

\item Entry 2. The model ensemble is carried out for both Task1 and Task2. In Task3, the response is generated using beam search, with a beam size of 5.

\item Entry 3. The model ensemble for Task1 and Task2 is the same as entry 2. In Task3, the response is generated using beam search, with a beam size of 3.

\item Entry 4. The model ensemble for Task1 and Task2 is the same as entry 2. In Task3, the response is extracted directly from the retrieved knowledge snippet.
\end{itemize}
For the Task1 model of entry 0, it can be represented as SOP-32L-Context in short, where SOP-32L refers to the pre-trained 32L evaluation model optimized with SOP loss and the knowledge-seeking turn detection is estimated based on the dialogue context. Similarly, for the Task1 model of entry 1, it is referred as SOP-32L-Schema. For the Task2 model of entry 0-1, it is referred as SOP-32L-Selection. To carry out model ensemble, extra pre-training models are employed, including SOP-24L, NSP-24L, BERT-base \cite{devlin2019bert} and ALBERT-xlarge \cite{lan2019albert}.\footnotemark[1] 
\footnotetext[1]{The 24L models have 24 transformer blocks and 16 attention heads, with the hidden embedding dimension of 1024. Besides SOP, next-sentence-prediction (NSP) is another commonly used pre-training loss function. BERT and ALBERT are included for the sake of distribution diversity.}
For the Task1 of entry 2-4, the following component approaches are used for ensemble via majority voting: SOP-32L-Context, SOP-32L-Schema, SOP-24L-Context, SOP-24L-Schema, NSP-24L-Context, ALBERT-xlarge-Schema, BERT-base-Schema. For the Task2 of entry 2-4, the following approaches are combined to calculate the average selection probability: SOP-32L-Selection, SOP-32L-Schema, SOP-24L-Selection, NSP-24L-Selection, ALBERT-xlarge-Schema.\footnotemark[2]
\footnotetext[2]{These models are selected through experiments on the validation set, under the objective to maximize the knowledge selection metric Recall@1.}

Among these 5 entries, entry 1 is the full version of our proposed solution.\footnotemark[3] The comparison between entry 0 and entry 1 is to reflect the difference of schema guided knowledge decision over conventional dialogue context-based decision. The comparison between entry 1 and entry 2 is to exhibit the improvements brought by the model ensemble in Task1 and Task2. The last 3 entries are to examine the distinct strategies of response generation.
\footnotetext[3]{Our source code and trained models will be released at \url{https://github.com/PaddlePaddle/Knover}.}

The experimental results on the validation set and the test set are summarized in Table \ref{tab:valid} and Table \ref{tab:test}, with the highest value written in bold. The official baseline method is based on GPT-2 \cite{radford2019language}, using dialogue context for turn detection and top-p sampling for response generation \cite{holtzman2019curious}. There are 24 teams participated in the competition, and our team id is 19. Several interesting phenomena can be observed from these results. 1) The dialogue context-based knowledge decision works well on the validation set. In comparison, the schema guided knowledge decision obtains superior performance on the test set, demonstrating better generalization on the unseen conversations from new domains or locales. 2) The obvious gap between the baseline and the proposed method indicates that the negatives enhanced strategy brings a significant improvement in knowledge selection. 3) As for response generation, although directly selecting the knowledge snippets as the reply ensures information accuracy, it achieves inferior performance compared with generation based methods. 
\begin{table}
	\centering
	\includegraphics[width=0.48\textwidth]{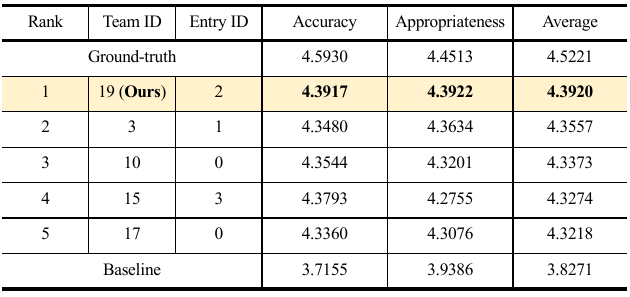}
	\caption{Final human evaluation on the test set, with the highest value written in bold.}
	\label{tab:human}
\end{table} 

\begin{figure}
	\centering
	\includegraphics[width=0.48\textwidth]{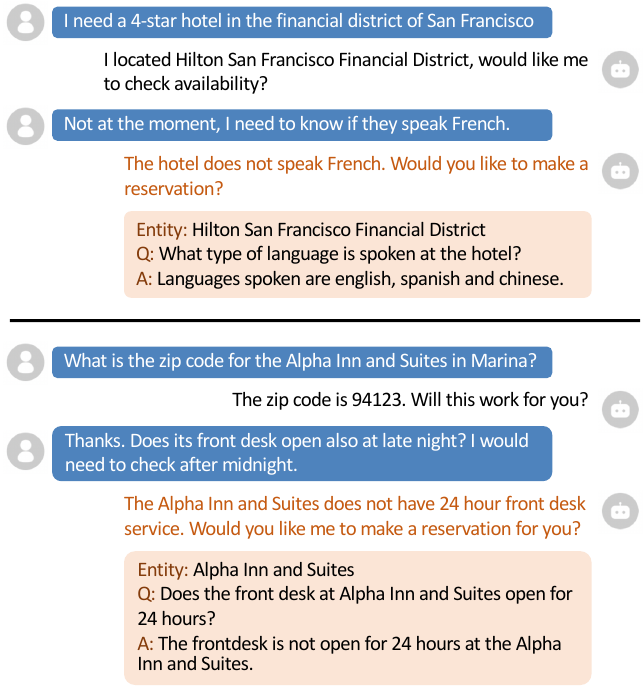}
	\caption{Cherry-picked examples.}
	\label{fig:good_case}
\end{figure} 
\begin{figure}
	\centering
	\includegraphics[width=0.48\textwidth]{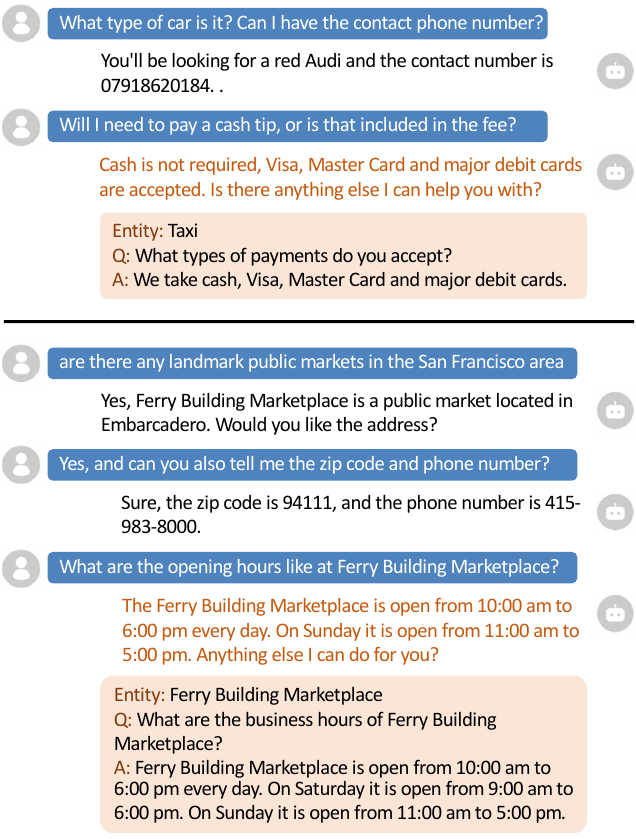}
	\caption{Examples with issues.}
	\label{fig:bad_case}
\end{figure} 
Besides the objective evaluation, human evaluation has been carried out for the final ranking. For the top 12 teams, the entry with the best objective evaluation results will be selected for the final human evaluation. Two metrics are considered in the evaluation: appropriateness and accuracy. Appropriateness measures how well the system response is naturally connected to the conversation. Accuracy accesses how accurate the system’s response is given the reference knowledge. The evaluation score ranges from 1 to 5. The higher, the better. The final evaluation results on the test set are summarized in Table \ref{tab:human}. Our proposed approach obtains 1st place in the final ranking. The golden responses are also enrolled in the final evaluation. The small gap between our approach and ground truth suggests that the system is able to provide high-quality and human-like services via external knowledge access. 

\subsection{Discussions}
To further analyze the quality of the proposed approach, several good and bad cases are provided in Figure \ref{fig:good_case} and Figure \ref{fig:bad_case}. From these cherry-picked examples shown in Figure \ref{fig:good_case}, it can be observed that the model is able to select the most appropriate knowledge snippet from the large-scale external database and generate high-quality knowledge grounded responses. In the upper case, the accurate answer towards the user's query is not included in the surface contents of the knowledge snippet. The model still generates an accurate reply, exhibiting the ability of natural language inference to some extent.

For those examples in Figure \ref{fig:bad_case}, the issues mainly come from two aspects. First, the model fails to select the most appropriate knowledge snippet due to the deficiency on training samples of some patterns. This issue might be alleviated through more advanced retrieval techniques or response generation conditioned on multiple knowledge snippets. Second, for those complicated knowledge snippets with multiple segments, sometimes the model omits a minor part and fails to generate complete information. This issue might be alleviated through the combination of extraction and generation in the near future.

\section{Related Work}
The related work will be discussed on task-oriented dialogue systems and knowledge-grounded response generation. 

Task-oriented dialogue systems interact with users in natural language and help them to accomplish some tasks, such as setting an alarm clock, booking a taxi, reserving a table, and so on. Conventional systems \cite{young2013pomdp, henderson2014second, wen2015semantically} adopt the modular architecture, including natural language understanding (NLU), dialogue state tracking (DST), dialogue policy, and natural language generation (NLG) modules. Recently, some end-to-end neural models \cite{wen2017network, li2017end, ham2020end} have been introduced for task-oriented dialogue systems. Regardless of the modular or end-to-end architecture, these systems need to operate within the scope of pre-defined APIs and cannot handle those out-of-range queries. Considering that relevant information about these queries might already exist on the internet, a new way is paved to incorporate external knowledge into task-oriented conversation modeling \cite{kim2020beyond}. 

To improve the informativeness in social conversations, some approaches \cite{dinan2018wizard, lian2019learning, fan2020augmenting} have explored knowledge grounded dialogue generation, where relevant knowledge segments are retrieved and encoded into memory network. As there exists the one-to-many mapping phenomenon in open-domain social conversations \cite{zhao2017learning, kim2019sequential, bao2019plato}, multiple knowledge segments might be appropriate to produce coherent responses. In comparison, task-oriented conversation modeling needs to deliver precise information to satisfy the user's needs. Therefore, it encounters more stringent requirements on knowledge selection and utilization. In this work, we have explored several advanced techniques to enhance task-oriented dialogue generation via external knowledge access. 

\section{Conclusion}
To boost the capacity of task-oriented dialogue system, in this work, we have explored several advanced techniques, including schema guided knowledge decision, negatives enhanced knowledge selection, and knowledge grounded response generation. Comprehensive experiments have been carried out on the publicly available dataset. Experimental results demonstrate that the schema guided knowledge decision achieves better generalization on unseen conversations and negatives enhanced knowledge selection brings significant improvements. More coherent and accurate knowledge grounded responses are generated by leveraging powerful pre-training models. As compared with other state-of-the-art methods, our approach obtains superior performance and ranks the 1st in the final evaluation.

\section*{Acknowledgments}
We would like to thank the reviewers for their constructive suggestions; Jingzhou He, and Tingting Li for the help on resource coordination; Wenquan Wu, and Han Zhou for the helpful discussions. This work was supported by the Natural Key Research and Development Project of China (No. 2018AAA0101900).

\bibliography{bibtex}
\end{document}